\documentclass[11pt]{article}

\usepackage[preprint]{acl}

\usepackage{times}
\usepackage{latexsym}

\usepackage[T1]{fontenc}

\usepackage[utf8]{inputenc}

\usepackage{microtype}

\usepackage{inconsolata}

\usepackage{graphicx}

\usepackage{amsmath}
\usepackage{amsfonts}
\usepackage{url}

\usepackage{balance}
\usepackage{multicol}
\usepackage{todonotes}

\title{\textsc{KnowledgeDebugger} -- an Exploration Tool for Knowledge Localization and Editing in Transformers}

\author{Eric Benz \and Lennart Stöpler \and Nikolai Bolik \and Artur Andrzejak\\
Heidelberg University \\
\texttt{\{eric.benz, lennart.stoepler, nikolai.bolik, artur.andrzejak\}@uni-heidelberg.de} \\}

\begin{document}
\maketitle
\begin{abstract}
Recent research has increasingly focused on understanding how Transformers store and process knowledge, as well as how this knowledge can be edited. Research work in this area is often conducted in two phases: first, phenomena are explored on individual samples. Then, when results appear promising, more statistically robust experiments follow.
 To support the first phase, we propose \mbox{\textsc{KnowledgeDebugger}}, a GUI-based exploration tool for knowledge localization and editing in Transformers. Our tool -- inspired by \mbox{\textsc{LM-Debugger}}~\citep{geva-etal-2022-lm} -- offers no-code access to the methods in \mbox{\textsc{EasyEdit}}~\citep{wang2023easyedit}, a widely used library of state-of-the-art Knowledge Editing approaches. We demonstrate the tool’s effectiveness through case studies of recent findings in this field.\footnote{The code is available at \url{https://github.com/ebnz/lm-debugger}}\footnote{The documentation is available at \url{https://ebnz.github.io/lm-debugger/}}
\end{abstract}
 
\section{Introduction}
\label{sec:intro}
Growing research effort is being invested in understanding the generative processes of large language models (LLMs) in terms of human-interpretable concepts (\citet{geva-etal-2021-transformer, elhage2022toymodelssuperposition, lindsey2025biology}; \textit{inter alia}). A complementary line of work seeks to edit a model’s knowledge directly on its parameters (\citet{meng2023locatingeditingfactualassociations, aging_hartvigsen_2022, precise_pan_2025}; \textit{inter alia}). Many of these Knowledge Editing (KE) methods have recently been collected in the \textsc{EasyEdit} library \citep{wang2023easyedit}. By unifying a diverse set of techniques within a common framework, \textsc{EasyEdit} simplifies their evaluation and mutual comparison.\\
The process of 
hypothesis formation in these domains greatly benefits from developing intuition about the phenomena that arise during the model’s forward pass—for example, changes induced by KE interventions. To this end, \textsc{LM-Debugger}~\citep{geva-etal-2022-lm} was proposed as a tool for interactive inspection of model activations down to the granularity of individual neurons. The tool further enables interventions on the model's output by modulating the contribution of specific neurons during the forward pass. However, it does not cover mainstream KE methods and does not offer to modify the underlying model parameters.\\
To close this gap, we propose \textsc{KnowledgeDebugger}, an interactive GUI-based tool for exploring the impact of state-of-the-art KE methods for Transformer architectures collected in the well-maintained \textsc{EasyEdit} library.
By building upon and extending the GUI interface of \textsc{LM-Debugger}, it offers interactive access to KE methods as well as pre-defined and user-defined metrics for evaluating the impact of KE interventions across all Transformer layers. Our tool makes it easier for newcomers to develop an intuition of available KE methods. Moreover, researchers already familiar with the state-of-the-art can obtain a qualitative impression of a method’s behavior before conducting more systematic evaluations. \\
This work makes the following contributions:
\begin{enumerate}
\item We propose a GUI-based tool for exploring  KE methods on Transformer models.
\item We equip our tool with a set of metrics tailored to evaluating KE methods, along with a low-barrier interface for defining custom metrics. In addition, we provide a flexible checkpointing system to facilitate reproducibility and accelerate workflow.
\item We demonstrate our tool on three case studies drawn from recent KE literature, confirming its effectiveness in investigating KE methods.
\end{enumerate}

\section{Background}
\label{sec:background}

\subsection{Theoretical Framework}
The field of mechanistic interpretability in general and KE in particular has coalesced around a particular interpretation of the Transformer architecture. We will briefly summarize the core ideas in this section and provide references for further reading.

\subsubsection{The Residual Stream}

In Transformer architectures, every block--Attention and Multi-Layer-Perceptron (MLP)--is bypassed by a residual connection \citep{attentionisallvaswani}. Recent work in mechanistic interpretability has highlighted the central functional importance of these residual connections and proposed to view them as the main vector pathway that carries the model’s internal representation across layers \citep{elhage2021mathematical}. In this interpretation the \textit{residual stream} (i.e., the collective of all residual connections) is updated at every block, as modules read from this shared representation and write their outputs back into it. We say \textit{residual stream (vector)} for the vector of activations at any point along this pathway.

\subsubsection{Unembeddings of the Residual Stream}
Crucially, any residual stream vector can be projected back into the vocabulary space by applying the model's unembedding matrix, which maps the activations to logits over the vocabulary \citep{nostalgebraist}. Thus, given a residual stream vector \(x\), the projection \(W_{\text{unemb}} \cdot x\) yields a score for every vocabulary token, corresponding to the logit value of that token under the current intermediate representation. Sorting these scores identifies the tokens most closely aligned with the current intermediate activations.

\subsubsection{The MLP as a Key-Value Store}
The MLP sublayer of a Transformer is defined as:
\begin{align}
    \operatorname{MLP}(x) := W_{\text{down}} \cdot f(W_{\text{up}}^\top\cdot x), 
\end{align}
where \(W_{\text{up}}\) and \(W_{\text{down}}\) are the MLP encoder and decoder matrices, respectivly, with shapes \(\mathbb{R}^{d \times d_m}\) and typically \(d_m > d\). Here, \(x \in \mathbb{R}^d\) ($d$ is the residual stream dimension), and \(f\) denotes a non-linear activation function.
An idea shared by many authors  is to interpret this construct as a key-value store. There are two complementary ways to do this. Proponents of the per-neuron view argue that columns of $W_{\text{up}}$ (each of which is in $\mathbb{R}^d$) constitute the keys and the columns of $W_{\text{down}}$ form the value vectors \citep{geva-etal-2021-transformer}. In this case the MLP is an element-wise mapping of keys to values.  
The per-layer interpretation instead posits that the intermediary activation $f(W_{\text{up}}^\top\cdot x)$ is the key and that $W_{\text{down}}$ maps this key to a value \citep{meng2023locatingeditingfactualassociations}. In this case there are infinitely many keys--mapped to infinitely many values.\footnote{Under a sparse neuron basis, such as proposed by \citet{transcodersdunefsky}, both views coincide.}

\subsection{Knowledge Editing}
\label{sec:knowledge_editing}
Knowledge Editing (\citet{meng2023locatingeditingfactualassociations, aging_hartvigsen_2022, precise_pan_2025}; \textit{inter alia})
attempts to change the parametric knowledge in LLMs by targeted edits on the model weights. The challenge is to change specific factual associations, while preserving the model’s overall behavior. Factual associations are typically framed in terms of knowledge triplets $(e_\text{head}, r, e_\text{tail})$, where $e_\text{head}$ is the head entity, $e_\text{tail}$ is the tail entity, and $r$ is the relation between them. An example would be ("Rome", "\{\} is located in", "Italy"). The goal is to find a transformation $\theta \longrightarrow \theta'$ of the model weights that encodes this knowledge triplet without changing any other aspect of the model's knowledge. The task is typically evaluated on the following terms: (i) \emph{specificity}, i.e., only the targeted fact should change; (ii) \emph{locality}, meaning that the edit should be confined to a small subset of parameters; (iii) \emph{generality}, meaning that the revised fact holds across rephrasings and varying contexts, and (iv) \emph{robustness}, meaning that the modification should not degrade the overall model quality.
A wide range of techniques have been proposed to approach these desired properties, 
in particular: (i) hyper-networks~\citep{editing_decao_2021, fast_mitchell_2021, calibrating_dong_2022}, (ii) external codebooks~\citep{aging_hartvigsen_2022, melo_yu_2023, 10.5555/3737916.3739619}, and (iii) locate-then-edit methods~\citep{meng2023locatingeditingfactualassociations, meng2023masseditingmemorytransformer, li2024pmetprecisemodelediting, precise_pan_2025, fang2025alphaeditnullspaceconstrainedknowledge}.
\subsection{\textsc{EasyEdit}}
\textsc{EasyEdit}\footnote{Available at \url{https://github.com/zjunlp/EasyEdit}} \citep{wang2023easyedit} collects many state-of-the-art KE methods 

in one library. Apart from serving these methods via a common and model-agnostic API it also provides a shared testing framework. \textsc{EasyEdit} is community-driven and actively maintained, ensuring that methods accessed through \mbox{\textsc{KnowledgeDebugger}} remain up-to-date. 
\subsection{\textsc{LM-Debugger}}
The \mbox{\textsc{LM-Debugger}}~\citep{geva-etal-2022-lm} is a tool to examine the model's behavior and for interacting with its internal computations at the level of MLP value vectors, i.e., columns of the MLP down matrix. It offers three core functionalities which are preserved within \mbox{\textsc{KnowledgeDebugger}} and  briefly discussed below.
\subsubsection{Trace View}
\label{sec:trace_view}
The Trace View displays, for each layer, the top-ten unembeddings of the residual stream into the vocabulary before and after the MLP update. This allows users to observe how the model’s internal state evolves across layers. It also highlights the ten most influential MLP value vectors, ranked by their activation value, showing the dominant contributions. Each vector can be inspected individually by projecting it into the vocabulary space and displaying the highest-scoring tokens under this projection. This functionality is fully preserved in our tool. The \textsc{TopKTokens} (Figure~\ref{fig:1} ~ B5) and the panel \mbox{\textsc{LMDebuggerIntervention}}
 (Figure~\ref{fig:1}~C8) expose the corresponding information.

\subsubsection{Explorer View}
\label{sec:explorer_view}
The Explorer View performs a static analysis of the model’s MLP value vectors by projecting each vector into the vocabulary space, producing token logits that indicate the tokens it most strongly promotes. \mbox{\textsc{LM-Debugger}} indexes these projections for all value vectors and thus enables a keyword-based search over their associated tokens. This allows users to quickly identify value vectors whose projections align with specific semantic domains. Users can access this functionality in our tool via the \mbox{\textsc{Explore}}-button, see Figure~\ref{fig:1}~A2.
\subsubsection{Interventions}
\label{sec:neuron_interventions}
\begin{figure*}[!t]
    \centering
    \includegraphics[width=\linewidth]{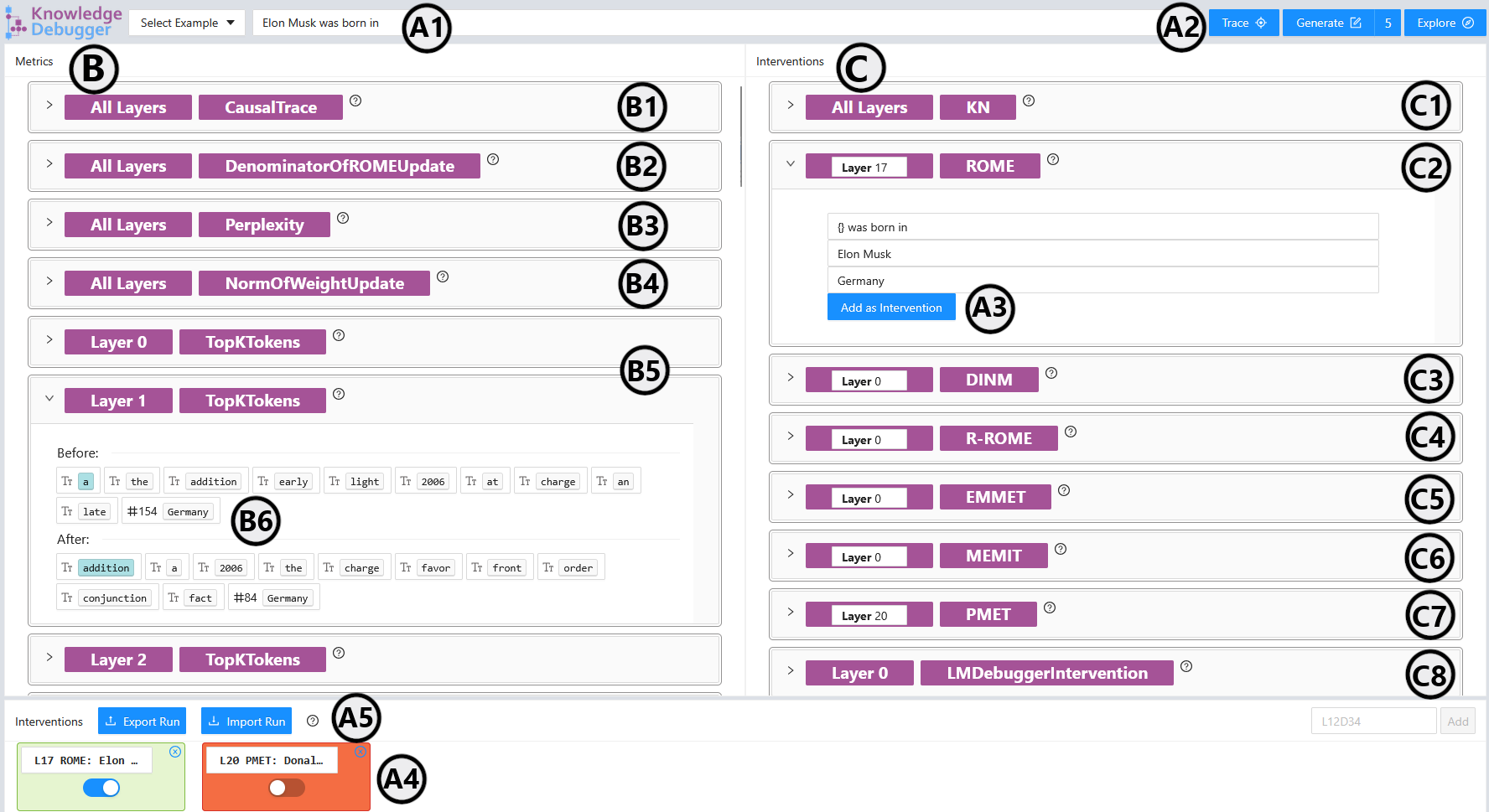}
    \caption{The figure illustrates the main functionality of \textsc{KnowledgeDebugger}. 
    Interface components are grouped into three categories: (A) general UI controls, (B) available metrics, and (C) intervention methods.\\ 
    (A) General UI controls:  
    (A1) The prompt input field.  
    (A2) Three main action buttons: \textsc{Trace}, \textsc{Generate}, and \textsc{Explore}. 
    \textsc{Trace} triggers the computation of the metrics and interventions and also displays available editing methods for the given prompt; 
    \textsc{Generate} produces the model's next-token predictions; 
    \textsc{Explore} is described in Section~\ref{sec:explorer_view}.  
    (A3) KE methods can be added to the intervention queue, provided they are correctly formatted.   
    (A4) The intervention queue, where left-to-right order determines the order of application. 
    The toggle enables or disables each method; for \textsc{LM-Debugger} interventions, the toggle behavior follows the description from Section~\ref{sec:neuron_interventions}.  
    (A5) The export and import functionality described in Section~\ref{sec:checkpointing}, allows runs to be saved and uploaded via a JSON file for fast reproducibility.\\ 
    (B) Metrics:  
    (B1)--(B4) The user metrics described in Section~\ref{UserMetrics}.  
    (B5) The top-ten token unembeddings, described in Section~\ref{sec:trace_view}.  
    (B6) The rank of the target token during a KE intervention, as described in Section~\ref{sec:rank_target}.\\ 
    (C) Intervention methods:  
    (C1)--(C7) knowledge-editing methods from the \textsc{EasyEdit} library.  
    (C8) \textsc{LM-Debugger} interventions, described in Section~\ref{sec:neuron_interventions}.
}
    \label{fig:1}
\end{figure*}

The \mbox{\textsc{LMDebuggerInterventions}} (Figure~\ref{fig:1}~C8) allow users to manipulate the MLP value vectors during inference. For any of the pre-selected top-ten value vectors, the user can apply an intervention, a binary switch that either suppresses this vector or amplifies its contribution. Turning off a vector sets its activation coefficient to zero, effectively removing its influence on the MLP update. Turning it on assigns the vector the same activation magnitude as the maximum amplitude in that layer, ensuring that it becomes one of the dominant contributors.

\section{KnowledgeDebugger}
\label{sec:KnowledgeDebugger}

\subsection{Knowledge Editing Interventions}

In addition to the neuron-level interventions supported by \textsc{LM-Debugger} and discussed in Section~\ref{sec:neuron_interventions}, \mbox{\textsc{KnowledgeDebugger}} incorporates various KE methods from the \textsc{EasyEdit} library~\citep{wang2023easyedit}, allowing the user to alter the LLM's knowledge directly on its weights. 
Users configure these interventions in the UI, add them to an intervention queue (Figure~\ref{fig:1}~A3) and arrange them in arbitrary order using the drag-and-drop menu (Figure~\ref{fig:1}~A4).

The queue order determines the order in which the interventions are executed.\\
The methods themselves are accessed through an adapter-class which wraps their underlying \textsc{EasyEdit} implementation. Currently, we have configurations available for 7 additional KE methods (Figure~\ref{fig:1}~C1--C7), among which are state-of-the-art methods such as ROME \citep{meng2023locatingeditingfactualassociations}, PMET \citep{li2024pmetprecisemodelediting}, and MEMIT \citep{meng2023masseditingmemorytransformer}. 

\subsection{Extended Metrics}

\subsubsection{Rank of target token}
\label{sec:rank_target}
In most KE scenarios the target tail entity \(e_\text{tail}\) plays a critical role. Analyzing its emergence in the residual stream can be instructive in understanding how knowledge is stored and processed. In particular, we would like to know its rank in the unembeddings even if that rank is not among the top ten. To this end, we track the rank of the target token for all edits, in addition to the top-ten unembeddings already discussed in Section~\ref{sec:trace_view}. 

\subsubsection{User-Defined Metrics}
\label{UserMetrics}
While the ranked token unembeddings are a powerful analytical tool, they potentially provide an incomplete view of the generation process. To better support researchers in tracking properties relevant to their specific research questions, we added an easy-to-extend interface for user-defined metrics.  Using this interface, the current version of our tool provides the following metrics:\\

\noindent \textbf{Causal Trace} (Figure~\ref{fig:1}~B1). 
The authors of \citet{meng2023locatingeditingfactualassociations} use causal tracing as a way to localize knowledge in the model weights. Assume that a model generates a knowledge triplet (\textit{subject}, \textit{relation}, \textit{object}), i.e., if we prompt the model with "\textit{subject} \textit{relation}" it will predict "\textit{object}". Causal tracing localizes the activations (i.e., token and layer) in this sequence which can be most strongly attributed to the prediction of "object". Methods such as ROME use this information to choose a layer to edit. \\

\noindent \textbf{Norm of Weights Update} (Figure~\ref{fig:1}~B4).
Given an MLP decoder matrix $W$, the ROME method inserts a new key-value pair $(k_\star, v_\star)$ by applying the following update~\cite{meng2023locatingeditingfactualassociations}:
\begin{align}
    \hat{W} &= W + \Delta,
        \label{eq:rome_W_update} \\
    \Delta &= (v_\star - W k_\star)\,
              \frac{(C^{-1} k_\star)^{\top}}
                   {(C^{-1} k_\star)^{\top} k_\star},
        \label{eq:rome_delta_update}
\end{align}
where \(k_\star \in \mathbb{R}^{d_m}\), \(v_\star\in\mathbb{R}^{d}\) and \(C = K K^{T}\) is the (uncentered) covariance matrix of keys sampled from a representative corpus (i.e., \(K \in \mathbb{R}^{d_m \times n}\) stacks \(n\) key vectors $k_i$ extracted from the target layer across a set of contexts).

We quantify the strength of a ROME update as the Frobenius norm of the update matrix \(|\Delta|\). Recent work \citep{gupta2024rebuildingromeresolving, yang-etal-2024-fall} has identified excessively large weight updates as a cause of model collapse.\\

\noindent \textbf{Denominator of ROME Update} (Figure~\ref{fig:1}~B2). The work of \citet{yang-etal-2024-fall} further identifies the denominator in Equation~\eqref{eq:rome_delta_update} as the source of excessively large updates and consequent model collapse, making it an additional indicator of problematic edits.\\

\noindent \textbf{Perplexity} (Figure~\ref{fig:1}~B3).
In other work, the model's perplexity on a small held-out dataset is proposed as an indicator of model degradation \citep{yang-etal-2024-butterfly}. The authors show that a decline in general language modeling capabilities--as indicated by high perplexity--strongly correlates with poor downstream performance. In line with their work we report the average perplexity on the ME-PPL50 dataset.

\subsection{Checkpointing}
\label{sec:checkpointing}
To boost the reproducibility of different runs, \mbox{\textsc{KnowledgeDebugger}} introduces a checkpointing mechanism for experimental configurations. These configurations can be exported to a JSON-file and then be shared or imported again into the application (Figure~\ref{fig:1}~A5). Configuration exports contain all parameters of a run, along with the run’s outcome. For a quick introduction to \mbox{\textsc{KnowledgeDebugger}}, we provide exports of the scenarios in Section~\ref{sec:case_studies} via the application's repository and via a shared cloud folder.\footnote{The exported scenarios can be downloaded at \url{https://heibox.uni-heidelberg.de/d/d665298efa304dce95b5/}}\looseness=-1
\subsection{Model Selection}
\label{sec:model_selection}
\mbox{\textsc{KnowledgeDebugger}} supports all Transformer-based model architectures.\footnote{Individual KE methods might place further restrictions on the choice of model. This is a limitation inherited from those methods and their implementation in \textsc{EasyEdit}, not a limitation of this tool.} Model selection is handled via a lightweight backend setup step by specifying the path to a model-specific configuration file. We provide configuration presets for several widely used Transformer models. 

Details on the configuration are documented in the project's repository.

\section{Case Studies}
\label{sec:case_studies}

To showcase the practical utility of our tool, we present three case studies -- all performed on \textsc{GPT2-XL}~\citep{radford2019language}, a model that remains commonly used in contemporary interpretability work \citep{fang2025alphaeditnullspaceconstrainedknowledge, cohen-etal-2024-evaluating, gupta-etal-2024-model} and runs on most hardware architectures, thus facilitating reproducibility of the case studies for a broader audience. To demonstrate that \mbox{\textsc{KnowledgeDebugger}} scales to substantially larger architectures, we include an additional case study using Llama-3.1 in the Appendix \ref{sec:appendix}.
We recommend downloading the provided checkpoint files (Section \ref{sec:checkpointing}) to follow along interactively.\footnote{A live demo of the tool is available at \url{https://knowledgedebugger.com/}} To simplify notation, all hints refer to Figure~\ref{fig:1} (e.g., (B5) means Figure~\ref{fig:1}, element~B5).\footnote{A video illustrating the first case study is available at \url{https://youtu.be/PXsEC_KfHiU}}
\subsection{Understanding unnatural edits}
The ROME-family of KE methods is known to cause disruptive edits~\cite{nanda2024opinionatedlist, hoelscher-obermaier-etal-2023-detecting, cohen-etal-2024-evaluating, gupta-etal-2024-model, yang-etal-2024-fall}. 
Our tool provides good intuition for how ROME-injected knowledge differs from pre-edit parametric knowledge.

To illustrate this, we analyze the following knowledge triplet: (“Elon Musk”, “\{\} was born in”, “South Africa”) and attempt to edit the model to instead predict (“Elon Musk”, “\{\} was born in”, “Germany”). First, we prompt the unedited model with the sentence "Elon Musk was born in" (A1). As expected, the model continues the sentence with "South Africa". We observe that an association with South Africa first emerges in the top-ten token unembeddings (B5) at layer 39. Here, the 8th most likely token is "Cape"--likely a prefix of the city of Cape Town. The token "South" emerges three layers later as the top most likely continuation. \\
Now we apply a ROME edit (C2 to create and A4 to enable the edit) at layer 17 (the EasyEdit default value\footnote{A causal trace (B1) would indicate layer 7 as the layer to edit. The results for editing this earlier layer are qualitatively similar.}). 
The model now successfully continues the prompt with “Germany.” However, the unembeddings indicate that the model treats the injected knowledge of "Germany" in a markedly different way: “Germany” becomes the top token as early as layer 24. If the edit had--as intended in KE--updated the model's parametric knowledge of Elon Musk's birthplace, we would expect the association to be processed similarly and to emerge at a comparable depth. Its emergence 15 layers earlier suggests this is not the case.

\subsection{Identifying collapse cases}
Another strength of \mbox{\textsc{KnowledgeDebugger}} is its ability to quickly test the robustness of findings from other publications. 

Multiple authors~\citep{gupta-etal-2024-model, gupta2024rebuildingromeresolving, yang-etal-2024-fall, yang-etal-2024-butterfly} 
have pointed out that in some cases even a single ROME edit can cause model collapse. \citet{yang-etal-2024-fall} observe this phenomenon for edits where the subject a) consists of a single token and b) is located at the very beginning of the prompt. These edits can be identified by a larger than normal ratio between numerator and denominator in the ROME update formula, Equations~\eqref{eq:rome_W_update} and~\eqref{eq:rome_delta_update}.
To demonstrate our framework, we implement the metrics used in these papers (cf. Section \ref{UserMetrics}) and reproduce their results in our tool. We analyze three cases--one that causes model collapse and two that do not. None of these cases were part of the samples analyzed in previous work. 
First, consider the edit ("Sri Lanka", "\{\} is located to the south of", "Australia").

The subject in this case is located at the beginning of the prompt, but consists of more than one token. In line with literature we observe that the perplexity (B3) of the model pre and post edit is nearly identical and both the norm of the update matrix (B4) and the absolute value of the denominator (B2) fall within the expected range\footnote{For non-collapse cases both values should be around 10.}.
Next consider this second edit: ("Germany", "The capital of \{\} is", "Paris").

This time the subject consists of only one token, but it is located in the middle of the prompt. As before, all metrics are nominal.
Finally, we edit the model as follows: ("China", "\{\} is located on the continent of", "Antarctica"). As expected, this single-token subject at the beginning of the prompt causes model collapse. We observe a $50$x increase in model perplexity. The denominator and the norm of the weight update deviate from their non-collapse ranges by three orders of magnitude\footnote{A fix to the ROME implementation that avoids these collapse cases has independently been proposed in \citet{gupta2024rebuildingromeresolving} and in \citet{yang-etal-2024-fall}. \citet{gupta2024rebuildingromeresolving}'s version of this fix (R-ROME) is included in the EasyEdit library and exposed in our tool. We verify that after this fix all metrics return to their non-collapse ranges for this edit.}.
We conclude that the published results generalize to unseen cases.

\subsection{Causual Tracing is a poor predictor for knowledge localization}

Furthermore, we illustrate the value of \mbox{\textsc{KnowledgeDebugger}} as a hypothesis-generation and exploratory research tool.
For the case of ROME-style KE methods, a stable knowledge localization is highly desirable, as model weights are edited at one specific layer. The \textsc{Causal Trace} metric (cf. Section~\ref{UserMetrics}) aims to identify where a model stores factual associations and thereby informs which layer to edit~\cite{meng2023locatingeditingfactualassociations}. If rephrasings of the same fact yield different layers, this choice becomes ambiguous. We evaluate the stability of this metric under reformulations of factual prompts. We selected prompts whose next-token prediction matched the intended tail entity $e_\text{tail}$ of the factual association.

For each valid prompt, we inserted the prefix (up to the tail entity) into a ROME intervention (C2), with the toggle disabled (A4). This triggers causal tracing (B1) without intervening in the models computational graph. 

Using this setup, we observed inconsistencies across rephrasings. For the triplet ("Eiffel Tower", "The \{\} is in the middle of", " Paris"), causal tracing localized the knowledge of "Paris" at layer \(16\), however for ("Eiffel Tower", “You can find the \{\} in the city of”) \(\rightarrow\) layer \(12\), and for ("Eiffel Tower", "The \{\} remains the most iconic structure in") \(\rightarrow\) layer \(20\).\\

Comparable variability appeared for the triplet (“Space Needle”, “The \{\} is located in downtown”, “Seattle”), where the knowledge of “Seattle” was traced to layer 18, whereas the rephrasing (“Space Needle”, “You can find the \{\} in downtown”, “Seattle”) traced to layer 14. These cases are also notable because the token "\_Seattle" already appeared in the top-ten unembeddings at layer \(3\) and \(2\) respectively, suggesting the information was present earlier than indicated by the trace.

This is likely due to an early-layer matching between “downtown” and "\_Seattle", as this behavior disappears in the case (“Space Needle”, “In the downtown district you will spot the iconic \{\} of”), which again traces to layer \(14\) but shows no early-layer appearance of “\_Seattle” in the unembeddings.\\
Further facts and rephrasings showed similar patterns. 
Although results tended to converge on a similar region of the model, the metric did not reliably pinpoint a single layer as the location of the fact. This variability would have to be investigated in depth to draw conclusive statements. However, these observations indicate that causal tracing may not be suitable for localizing associated knowledge within a LLM on the granularity of a single layer. This aligns with work by~\citet{NEURIPS2023_3927bbdc}, who showed that the layer most effective for edit success is largely unrelated to the layer identified by causal tracing.

\section{Conclusion}
\label{sec:conclusion}
We present \textsc{KnowledgeDebugger}, an interactive tool for the study of the behavior and outcomes of KE methods in Transformer models. 
Our evaluation demonstrated how it can be used to identify methodological shortcomings, enabled the efficient reproduction of prior results, and supported hypothesis generation for future research directions. We believe that \textsc{KnowledgeDebugger} will serve as a valuable resource for future work in the areas of interpretability and Knowledge Editing.

\bibliography{trimmed}
\clearpage
\appendix
\onecolumn
\section{Appendix}
\label{sec:appendix}
\begin{figure*}[h]
    \centering
    \includegraphics[width=\linewidth]{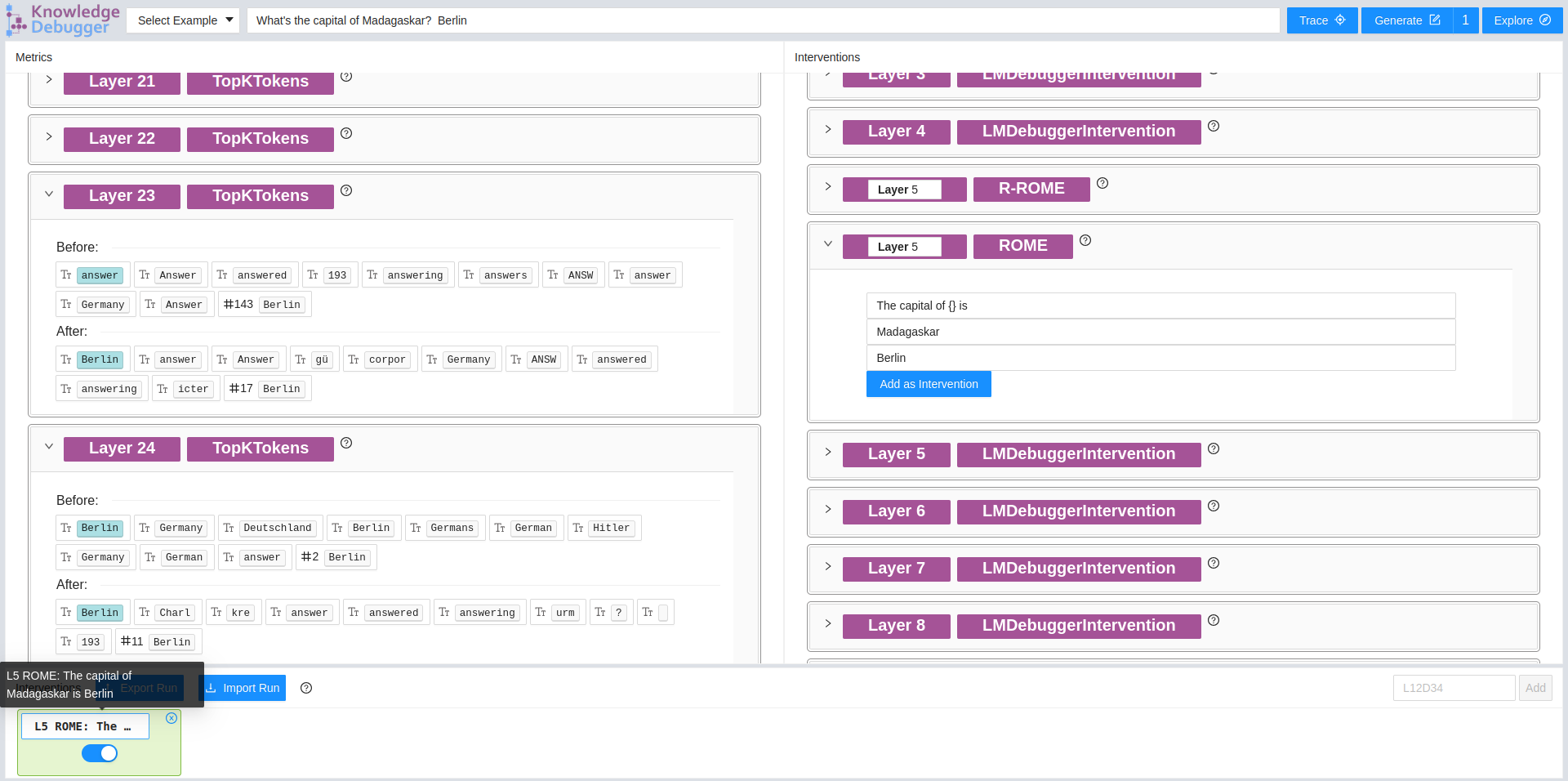}
    \caption{We apply a ROME update for the knowledge triplet ("Madagaskar", "The capital of \{\} is", "Berlin") at layer 5 of Llama~3.1 Instruct 8B. Starting at layer 23, "Berlin" becomes the top-ranked next-token prediction, and next token generation confirms that the model outputs "Berlin" as the next token, demonstrating a successful ROME intervention.
}
    \label{fig:2}
\end{figure*}
\begin{multicols}{1}
Figure~\ref{fig:2} illustrates a successful ROME edit applied at layer 5 using our tool. As shown, the target token Berlin becomes the highest-ranked prediction from layer 23 onward, indicating that the intended edit has been propagated through the subsequent layers. Notably, this edit was performed on Llama~3.1 Instruct 8B. In contrast, the other case studies in Section~\ref{sec:case_studies} were conducted on GPT2-XL to facilitate reproducibility on widely available hardware.
Since GPT2-XL has 1.5B parameters, the additional result in Figure~\ref{fig:2} further demonstrates that our tool is not tightly coupled to a single model scale; rather, its applicability generalizes across transformer-based language models.
\end{multicols}

\end{document}